\documentclass[runningheads]{llncs}

\usepackage{eccv}

\usepackage{eccvabbrv}

\usepackage{graphicx}
\usepackage{booktabs}

\usepackage[accsupp]{axessibility}  

\usepackage[pagebackref,breaklinks,colorlinks,citecolor=eccvblue]{hyperref}

\usepackage{orcidlink}

\def\sota{state-of-the-art}
\def\guidance{non-Lambertian surface regional guidance}

\def\aug{random tone-mapping augmentation}

\usepackage{multirow}
\usepackage{float}
\usepackage{marvosym}

\begin{document}

\title{Towards Robust Monocular Depth Estimation in Non-Lambertian Surfaces} 


\author{Junrui Zhang\inst{1}\orcidlink{0009-0004-8018-0458} \and
Jiaqi Li\inst{2,}\thanks{Corresponding author}  \orcidlink{0009-0004-7799-3407}  \and
Yachuan Huang\inst{2}\orcidlink{0009-0004-2314-3992} \and
Yiran Wang\inst{2}\orcidlink{0000-0002-2785-9638} \and \\
Jinghong Zheng\inst{2}\orcidlink{0009-0000-7996-8927} \and
Liao Shen\inst{2}\orcidlink{0000-0002-2423-4835} \and
Zhiguo Cao\inst{2}\orcidlink{0000-0002-9223-1863}}

\authorrunning{J.~Zhang et al.}

\institute{School of Future Technology, Huazhong University of Science and Technology \\ \email{jr\_z@hust.edu.cn} \and School of AIA, Huazhong University of Science and Technology\\ \email{\{lijiaqi\_mail,yachuan,wangyiran,deepzheng,leoshen,zgcao\}@hust.edu.cn}}

\maketitle

\begin{figure}
  \centering
  \includegraphics[height=6cm]{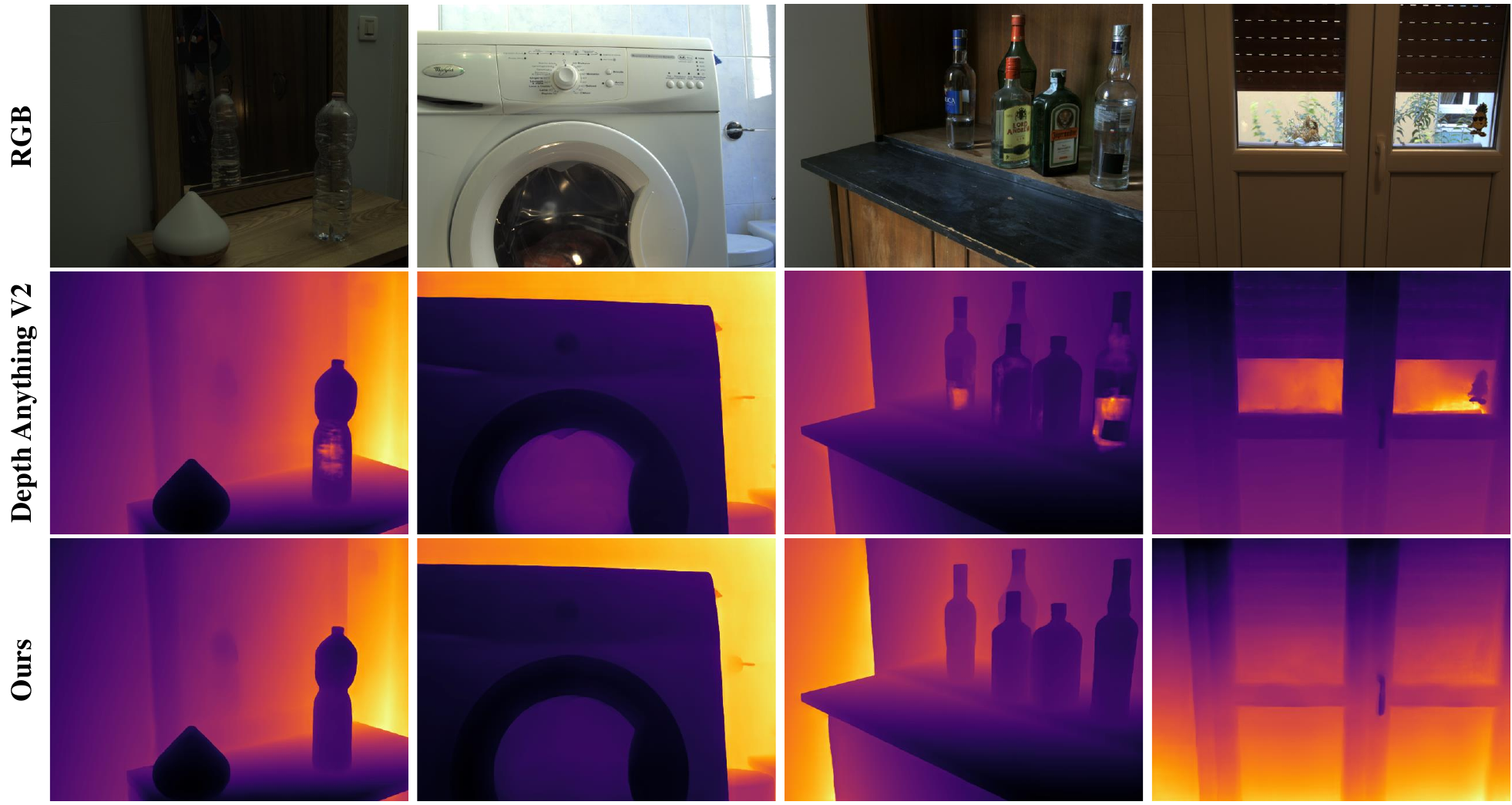}
  \caption{\textbf{Visual comparison of the baseline and our method.} Fine-tuned from Depth Anything V2, our method improves zero-shot performance in non-Lambertian. 
  }
  \label{fig:abstract}
\end{figure}

\begin{abstract}
  In the field of monocular depth estimation (MDE), many models with excellent zero-shot performance in general scenes emerge recently. However, these methods often fail in predicting non-Lambertian surfaces, such as transparent or mirror (ToM) surfaces, due to the unique reflective properties of these regions. Previous methods utilize externally provided ToM masks and aim to obtain correct depth maps through direct in-painting of RGB images. These methods highly depend on the accuracy of additional input masks, and the use of random colors during in-painting makes them insufficiently robust. We are committed to incrementally enabling the baseline model to directly learn the uniqueness of non-Lambertian surface regions for depth estimation through a well-designed training framework. Therefore, we propose non-Lambertian surface regional guidance, which constrains the predictions of MDE model from the gradient domain to enhance its robustness. Noting the significant impact of lighting on this task, we employ the random tone-mapping augmentation during training to ensure the network can predict correct results for varying lighting inputs. Additionally, we propose an optional novel lighting fusion module, which uses Variational Autoencoders to fuse multiple images and obtain the most advantageous input RGB image for depth estimation when multi-exposure images are available. Our method achieves accuracy improvements of 33.39\% and 5.21\% in zero-shot testing on the Booster and Mirror3D dataset for non-Lambertian surfaces, respectively, compared to the Depth Anything V2. The state-of-the-art performance of 90.75 in $\delta_{1.05}$ within the ToM regions on the TRICKY2024 competition test set demonstrates the effectiveness of our approach.
\end{abstract}

\section{Introduction}
In recent years, numerous approaches have been developed that significantly advanced the state-of-the-art in monocular depth estimation (MDE). These methods leverage semantic or surface normal information to enhance depth estimation accuracy \cite{depthfm,marigold,m3d2,da1} and utilize large-scale datasets to demonstrate robust zero-shot MDE capabilities \cite{da1,da2,m3d}. The rapid evolution in this field improves MDE model performance in traditional applications such as autonomous driving \cite{autodrive1} and 3D reconstruction \cite{3drecon1}, as well as in modern applications 
including novel view synthesis \cite{RF1,RF2} and 3D cinemagraphy \cite{li,shen}.

Despite the impressive performance of recent MDE models, estimating the depth of objects with non-Lambertian surfaces remains a big challenge for MDE models \cite{ntire2023,ntire2024,ignatov2022efficient}. 
Depth4ToM\cite{iccvToM} initially attempts to use an additionally required segmentation mask to in-paint the RGB with random colors, processes them with a pre-trained MDE model and aggregates the results to obtain a pseudo-labeled depth.
Even with many detailed designs, pseudo-labels that rely on the direct aggregation of network predictions are still not sufficiently reliable. Additionally, the random RGB in-painting requires extensive hyperparameter tuning, making the overall network structure heavily dependent on empirical adjustments. Therefore, we focus on enabling the network to directly learn the unique characteristics of non-Lambertian surfaces from the segmentation masks, thereby simplifying the pipeline and enhancing the stability of the training framework. Specifically, \guidance{} is designed to guide additional supervision during training, ensuring the network accurately estimates the depth of these special regions from gradient domain.

Compared to general depth estimation tasks, lighting or exposure conditions have a much greater impact on recovering the depth of non-Lambertian surfaces. Excessive lighting increases the dynamic range of the image, which can overwhelm the texture of transparent objects like glass, leading to erroneous depth predictions. At the same time, mirror reflections become more pronounced, causing the network to more easily predict the depth structure of the reflected content rather than the mirror surface. Conversely, insufficient lighting also hampers accurate depth estimation, introducing noise, blur, color distortion and drift, with point light sources having a more significant impact on the non-Lambertian surfaces and overall imaging. Therefore, we use \aug{} designed through experimentation to improve the robustness of the depth estimation network under various lighting conditions during training.

Moreover, using multiple exposure images of the same scene are shown to be generally beneficial for weakly-textured, transparent, or mirror scenes \cite{booster,booster2}. To adaptively fuse images with various exposure parameters for optimal prediction results, we innovatively employ Variational Autoencoders for fusion of RGB images before depth estimation.

Our method significantly improves the accuracy of the baseline model on non-Lambertian surfaces, achieving \sota{} 0.859 and 0.606 of $\delta_{1.05}$ in ToM regions on the Booster dataset and NYU Depth Dataset V2 of Mirror3D, respectively. Additionally, by using only \aug{} and \guidance{}, we achieve the best accuracy in the ToM regions on the TRICKY2024 competition test set \cite{2024tricky}. We also demonstrate the effectiveness of each proposed module in Sec. \ref{section:ablation}.


In summary, our contributions are as follows:
\begin{itemize}
\item Random tone-mapping augmentation is proposed to obtain a richer dataset to enhance the robustness of the MDE model, improving the zero-shot generalization capability for various lighting conditions.
\item We use \guidance{} to maintain the consistency of the ground truth and the depth predicted to improve the performance of the model on estimating the depth of non-Lambertian surfaces.
\item We propose an optional image fusion module to fully utilize multiple exposure information of the same scene, leveraging the prior knowledge of VAE.
\end{itemize}

\section{Related Works}

\subsection{Zero-shot Monocular Depth Estimation}
Recently the demand for zero-shot metric depth estimation is becoming increasingly important due to the needs of downstream tasks such as autonomous driving. Zoedepth~\cite{zoedepth} firstly proposes to use multiple lightweight heads to fit diverse data distributions from indoors to outdoors. By embedding camera priors into the network, ZeroDepth~\cite{zerodepth} achieves the best performance on autonomous driving datasets. Metric3D~\cite{m3d,m3d2} proposes a canonical and de-canonical camera transformation method to solve the metric depth ambiguity problems from various camera settings, improving the zero-shot generalization capability.

Due to the high cost of depth annotation, instead of predicting metric depth, most previous MDE methods~\cite{midas,dpt,kexian2018,megadepth} produce affine-invariant depth, \ie, the depth is up to an unknown scale and shift, to enable the model to generalize to unseen images after being trained on a few small-scale datasets.
Early works~\cite{megadepth,kexian2018} initially propose using multi-view data collected by large-scale networks and scale-invariant loss for training to enhance the generalization capability of MDE models. MiDaS~\cite{midas} further expands the natural scene dataset using 3D movies and achieves impressive generalization capability through mixed datasets and an improved scale- and shift-invariant loss function. DPT~\cite{dpt} builds upon this by improving the network architecture, introducing transformers~\cite{vaswani2017attention,wang2023neural} to increase network capacity and enhance performance. Recently, Depth Anything~\cite{da1,da2} achieves state-of-the-art zero-shot capability and performance by expanding the training dataset to tens of millions and replacing the encoder with Dinov2~\cite{dinov2}. Some methods~\cite{dadp,marigold,depthfm} achieve excellent depth edge quality and semantic integrity in zero-shot performance by introducing and improving diffusion models~\cite{ddpm}.

All of these methods focus on increasing the training data or improving the network structure but often neglect the fact that the majority of models fail to predict non-Lambertian surfaces, thereby limiting their practical applications. As stated in previous work \cite{iccvToM}, it is necessary to make targeted adjustments to the current MDE training framework for non-Lambertian surfaces. Our approach significantly enhances the robustness of existing models to non-Lambertian surfaces through \aug{} and specially designed \guidance{}.

\subsection{Dense Prediction of Non-Lambertian Surfaces}
For semantic segmentation, even though non-Lambertian regions and other regions differ greatly in RGB images, their manual annotation difficulty is almost the same. Until now, depth annotation methods represented by structured light, LiDAR, and ordinary stereo matching are still not suitable for non-Lambertian surfaces. As a result, there are more works on semantic segmentation for non-Lambertian surfaces. SATNet \cite{mirrordetection1} proposes a novel dual-path symmetry-aware transformer-based mirror segmentation network to learn symmetry relations for mirror detection, and PDNet \cite{depthawaremirror2} utilizes both RGB and depth and fuses different sizes of RGB features and depth features dynamically to outperform existing mirror segmentation models.

Different from the task of semantic segmentation, there are only a few works about monocular depth estimation on non-Lambertian surfaces. Mirror3D \cite{mirror3d} provides mirror annotations by humans for three RGBD datasets, which are NYU Depth Dataset V2 \cite{nyudv2}, Matterport3D dataset \cite{mp3d} and ScanNet dataset \cite{scannet}, and presents an architecture that predict the mirror mask to refine raw sensor depth or the output of MDE models. 
However, Depth4ToM \cite{iccvToM} uses RGB and non-Lambertian surfaces mask as the input. It in-paints the RGB images based on non-Lambertian masks with random colors and aggregates the outputs of a pre-trained MDE model from them, obtaining a pseudo-label for training the MDE model.

Compared with two methods mentioned above, we use large scale synthetic dataset \cite{Hypersim} which has nearly 77,400 images with highly precise depth annotations. Moreover, we directly train the MDE model to refine the depth in the non-Lambertian surface region, different from Mirror3D \cite{mirror3d}. Thus our method utilizes the semantic information of the non-Lambertian surfaces more concisely and effectively to guide the estimation of the depth in the non-Lambertian surface region instead of RGB in-painting in Depth4ToM \cite{iccvToM}. Specifically, we design a \guidance{} to guide the depth in the non-Lambertian surface region estimated by the MDE model to have the same consistent plane as the ground truth.

\section{Method}

\begin{figure}
  \centering
  \includegraphics[height=5.5cm]{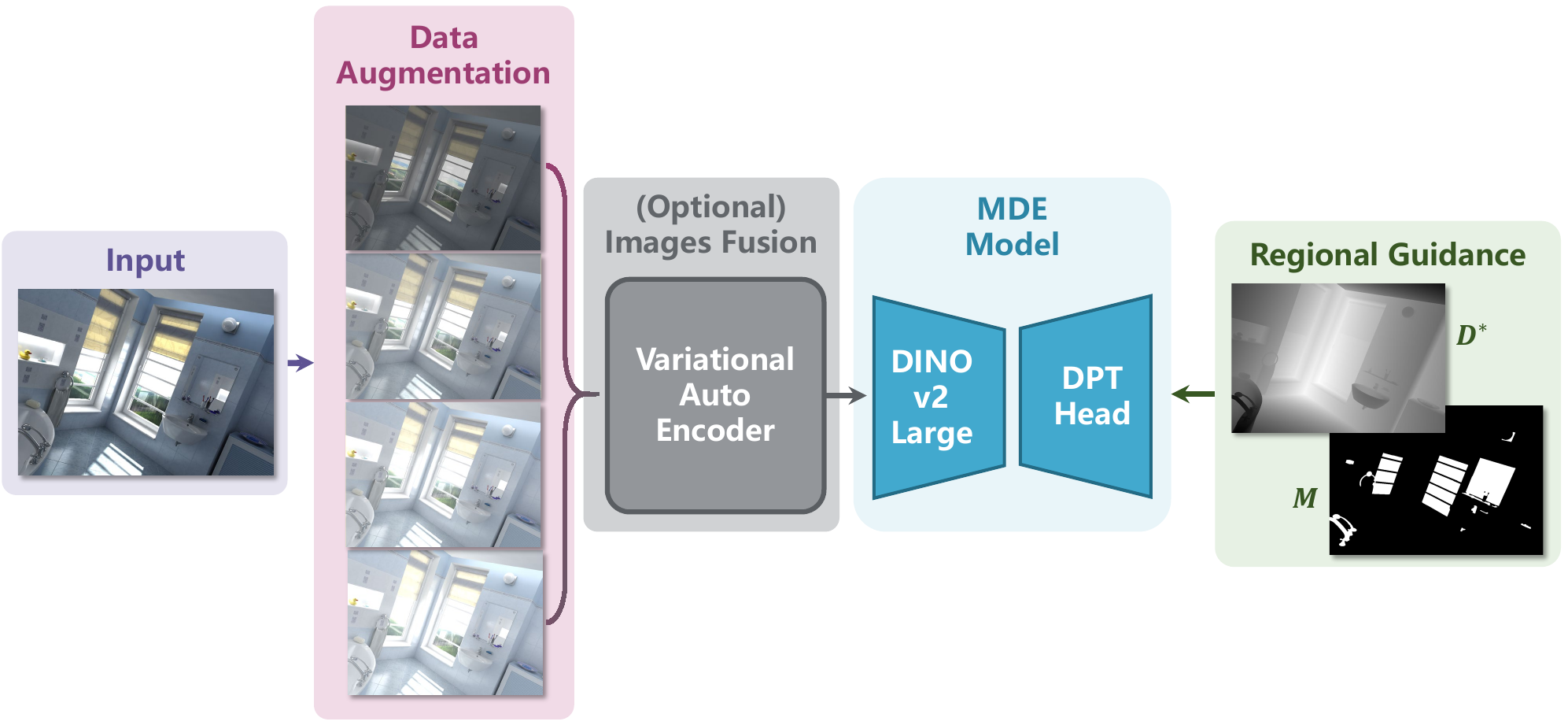}
  \caption{\textbf{Overview of our pipeline.} We train the large version of Depth Anything V2 on the Hypersim dataset with \aug{}. Non-Lambertian regional guidance computes and optimizes the loss in the non-Lambertian regions $M$ between the depth predicted and the ground truth $D^\star$ from the gradient domain. Images fusion is optional in inferring phase in case multi-exposure images are available.
  }
  \label{fig:pipeline}
\end{figure}

\subsection{Random Tone-mapping Augmentation}
As illustrated in these previous works \cite{booster,booster2,cui2023light}, lighting conditions have a much larger effect on depth estimation of non-Lambertian surfaces compared to ordinary scenes. To address this issue, we adopt the \aug{} with the aim to obtain a strong zero-shot depth estimation model even in the extreme lighting conditions. 

Specifically, since the Hypersim dataset \cite{Hypersim} is a synthetic dataset with all original digital assets are available, we can simply apply random tone-mapping to obtain a richer dataset with varying lighting condition during the training phase. The image $I_{aug}$ used to train MDE model after tone-mapping is obtained from the original image $I_{ori}$ by applying
\begin{equation}
I_{aug}=\alpha I_{ori}^{\gamma}
\end{equation}
where $\gamma = \frac{1}{2.2}$ is a standard gamma correction factor and $\alpha$ is a scale factor computed such that certain percentile intensity value maps to the value 0.8. For example, the $90_{th}$ percentile intensity value is $r_{90}$.

To augment the original images with different lighting conditions, we randomly adjust the percentile intensity value to a random value between 70 and 99. With different percentile intensity values, we can compute different scale factors and then get augmented images with different lighting conditions. 

\subsection{Non-Lambertian Surface Regional Guidance}
Our \guidance{} is based on an intuitive prior that the vast majority of ToM objects are continuous planes. To ensure that the network predicts the correct depth plane rather than the erroneous depth structure caused by reflections, we aim to constrain the predicted target region in the gradient domain to be as smooth as possible, consistent with the ground truth, in order to achieve the optimization objective.

Specifically, let $D$ and $D^\star$ denote the predicted depth map and the ground truth (GT) depth map, respectively. Let $\nabla$ represent the gradient magnitude. Given the non-Lambertian surfaces mask $M$, the regional gradient matching loss can be expressed as:
\begin{equation}
\begin{gathered}
\mathcal{L}_{ToM}=\frac{1}{M}\sum_{i=1}^M\Vert s \nabla D_i +t -\nabla D_i^\star \Vert_1 \,, \\
(s,t)=\mathop{\arg\min}\limits_{s,t}\sum_{i=1}^M\Vert (s \nabla D_i +t) -\nabla D_i^\star \Vert_2^2 \,,
\end{gathered}
\end{equation}
where $s$ and $t$ are the least squares coefficients. The $M$ for the Hypersim dataset is obtained by thresholding and binarizing the reflectance coefficient map. 

Since the mean-squared error is not robust to some outliers, the analytical solution of $s$ and $t$ using the least squares method will be greatly affected by outliers. To improve training, we define an alternative, robust loss function based on robust estimators of scale and shift following MiDaS \cite{midas}:
\begin{equation}
t(\nabla D)= median (\nabla D), s(\nabla D)=\frac{1}{M}\sum_{i=1}^M\vert \nabla D - t(\nabla D) \vert \,.
\end{equation}

We align both the $\nabla D$ and the $\nabla D^\star$ to have zero translation and unit scale:
\begin{equation}
\hat{\nabla D}=\frac{\nabla D - t(\nabla D)}{s(\nabla D)},\hat{\nabla D^\star}=\frac{\nabla D^\star - t(\nabla D^\star)}{s(\nabla D^\star)} \,.
\end{equation}

Then the non-Lambertian surface regional guidance loss is designed by trimming the $20\%$ largest residuals:

\begin{equation}
\mathcal{L}_{ToM}=\frac{1}{2M}\sum_{i=1}^{0.8M}\Vert \hat{\nabla D_i} - \hat{\nabla D_i^\star} \Vert_1 \,.
\end{equation}

\begin{figure}[ht]
  \centering
  \includegraphics[height=6cm]{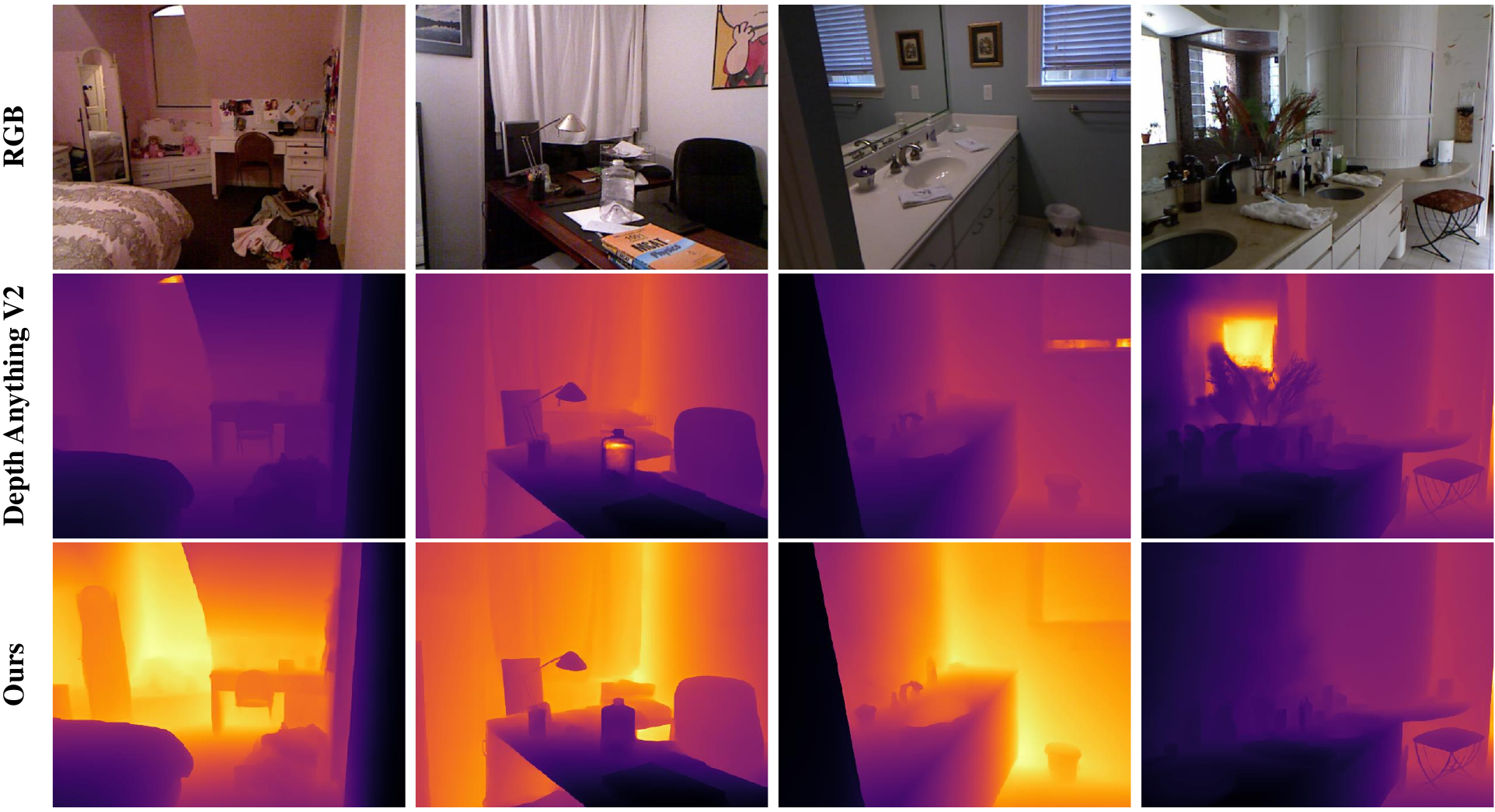}
  \caption{\textbf{Visual comparison of the baseline and our method.} The first row consists of images from the NYU Depth Dataset V2. The second and third rows are the monocular depth estimation results from the baseline and our method, respectively. Our method outperforms the baseline as we explicitly and directly use the semtanics of non-Lambertian regions to guide the depth estimation.
  }
  \label{fig:nyu}
\end{figure}
\subsection{Diverse Lighting Images Fusion}
Although images with \aug{} helps the model learn how to estimate the depth better during the training phase, the model may fail in some cases with extreme lighting conditions. In order to estimate the best depth for the scene, we propose an optional images fusion to process the images with the same scene but various lighting to a single image probably with the best lighting conditions for the depth estimation.

Specifically, we first stack the images with various lighting conditions but the same scene contents. Then the pre-trained  VAE provided by the Stable Diffusion \cite{SD} is used to encode each of the images and get the latent features. Finally, we aggregate the latent features and decode the obtained features via the decoder of the VAE to get a single image. Specifically, we process each image in the scene and get different latent features. Then we compute the mean of them to get the final features to be decoded.

Since the VAE provided by the Stable Diffusion is trained on KL-divergence loss for the image reconstruction using Open Images Dataset \cite{openimages} which is a dataset of  about 9M images annotated with image-level labels, object bounding boxes, object segmentation masks, visual relationships and localized narratives, the pre-trained VAE contains abundant semantic priors so that it can fuse the images with various lighting conditions to a single image which possesses the best lighting conditions for semantics.

\begin{table}
    \setlength{\tabcolsep}{4pt}
    \centering
    \caption{Quantitative zero-shot evaluations on the small NYU Depth Dataset V2. We compute the results in ToM regions by using the mirror masks which are annotated by Mirror3D. Our method outperforms the baselines in most cases. The best results in ToM regions are in \textbf{bold} and the second best results are \underline{underlined}.}
    \label{tab:zero_shot_nyu}
    \begin{tabular}{lccccccc}
        \toprule
        Method                             & Region & $\delta_{1.05}\uparrow$&  $\delta_{1.15}\uparrow$&  $\delta_{1.25}\uparrow$&  Abs Rel.$\downarrow$&  RMSE$\downarrow$&Log MAE$\downarrow$     \\
        \midrule
        \multirow{3}{2cm}{Depth Anything V1} & All    & 0.330 & 0.704 & 0.851                             & 0.131 & 0.361 & 0.042 \\
                                   & ToM    & 0.207 & 0.463 & 0.625 & 0.260 & 0.516 & 0.093 \\
                                   & Other  & 0.347 & 0.735 & 0.881 & 0.113 & 0.332 & 0.037 \\
        \midrule
        \multirow{3}{2cm}{Metric3D} & All    & 0.507 & 0.816 & 0.910                             & 0.089 & 0.266 & 0.037 \\
                                   & ToM    & 0.357 & 0.736 & 0.877 & 0.124 & 0.281 & 0.048 \\
                                   & Other  & 0.528 & 0.827 & 0.917 & 0.083 & 0.256 & 0.035 \\
        \midrule
        \multirow{3}{2cm}{DepthFM} & All    & 0.456 & 0.803 & 0.905                             & 0.094 & 0.286 & 0.040 \\
                                   & ToM    & 0.436 & 0.774 & 0.860 & 0.120 & 0.255 & 0.046 \\
                                   & Other  & 0.460 & 0.811 & 0.913 & 0.089 & 0.280 & 0.039 \\
        \midrule
        \multirow{3}{2cm}{Depth Anything V2} & All    & 0.632 & 0.901 & 0.951                             & 0.063 & 0.241 & 0.027 \\
                                   & ToM    & \underline{0.576} & \underline{0.919} & \textbf{0.962} & \underline{0.069} & \underline{0.158} & \underline{0.028} \\
                                   & Other  & 0.637 & 0.900 & 0.951 & 0.062 & 0.214 & 0.027 \\
        \midrule
        \multirow{3}{*}{Ours}      & All    &0.650       &0.915       &0.958       &0.060                                   &0.208       &0.025       \\
                                   & ToM    &\textbf{0.606}       &\textbf{0.929}       &\underline{0.960}       &\textbf{0.065}       &\textbf{0.147}       &\textbf{0.026}       \\
                                   & Other  &0.655       &0.913       &0.958       &0.059       &0.209       &0.025      \\
        \bottomrule
    \end{tabular}
\end{table}

\section{Experiments}

\subsection{Dataset and Metrics}
To evaluate the zero-shot capability of our method, we conduct zero-shot evaluations on the Booster dataset \cite{booster} and parts of NYU Depth Dataset V2\cite{nyudv2} which are refined by Mirror3D\cite{mirror3d}. Booster is a novel high-resolution, which is 4112 in width and 3008 in length, and challenging dataset with framing indoor scenes annotated by dense and accurate ground-truth disparities and several specular and transparent surfaces masks. We use the training dataset of the Booster which offers 456 samples with ground truth to evaluate the qualitative and quantitative results and use the test dataset of the Booster which offers 382 samples without ground truth to evaluate the qualitative results. We also conduct zero-shot evaluations on the small NYU Depth Dataset V2 of Mirror3D which contains 125 samples with a resolution of $456 \times 608$ for comprehensive comparison.

Due to the commonly used accuracy metrics, the $\delta_{2}$ accuracy and the $\delta_{3}$ accuracy, are both close to $1.0$ in the experiments, we choose accuracy metrics with lower thresholds to show the comparison more intuitively. The evaluation metrics are the accuracy $\delta_{1.05}$, $\delta_{1.15}$, and $\delta_{1.25}$, the Absolute Mean Relative Error (Abs Rel.), the Root Mean Square Error (RMSE) and the Mean Absolute Logarithmic Error (Log MAE) in all regions, the ToM regions and the other regions separately.

\begin{table}[ht]
    \setlength{\tabcolsep}{4pt}
    \centering
    \caption{Quantitative zero-shot evaluations on the Booster dataset. Our method, which is trained using the scale-shift-invariant loss and the ToM regional guidance, outperforms the baselines in most cases. The best results in ToM regions are in \textbf{bold} and the second best results are \underline{underlined}.}
    \label{tab:zero_shot_booster}
    \begin{tabular}{lccccccc}
        \toprule
        Method                             & Region & $\delta_{1.05}\uparrow$&  $\delta_{1.15}\uparrow$&  $\delta_{1.25}\uparrow$&  Abs Rel.$\downarrow$&  RMSE$\downarrow$&Log MAE$\downarrow$     \\
        \midrule
        \multirow{3}{2cm}{Depth Anything V1} & All    & 0.588 & 0.900 & 0.966                             & 0.061 & 0.052 & 0.026 \\
                                   & ToM    & 0.266 & 0.617 & 0.804 & 0.161 & 0.096 & 0.061 \\
                                   & Other  & 0.623 & 0.926 & 0.979 & 0.052 & 0.044 & 0.023 \\
        \midrule
        \multirow{3}{2cm}{Metric3D} & All    & 0.537 & 0.839 & 0.930                            & 0.078 & 0.068 & 0.033 \\
                                   & ToM    & 0.296 & 0.645 & 0.817 & 0.148 & 0.091 & 0.057 \\
                                   & Other  & 0.548 & 0.846 & 0.933 & 0.073 & 0.065 & 0.032 \\
        \midrule
        \multirow{3}{2cm}{DepthFM} & All    & 0.653 & 0.926 & 0.979                             & 0.051 & 0.045 & 0.022 \\
                                   & ToM    & 0.492 & 0.870 & 0.937 & 0.083 & 0.052 & 0.034 \\
                                   & Other  & 0.669 & 0.935 & 0.986 & 0.047 & 0.042 & 0.021 \\
        \midrule
        \multirow{3}{2cm}{Depth Anything V2} & All    & 0.802 & 0.975 & 0.993                             & 0.034 & 0.031 & 0.014 \\
                                   & ToM    & \underline{0.644} & \underline{0.869} & \underline{0.937} & \underline{0.081} & \underline{0.060} & \underline{0.031} \\
                                   & Other  & 0.830 & 0.988 & 0.998 & 0.029 & 0.025 & 0.013 \\
        \midrule
        \multirow{3}{*}{Ours}      & All    &0.890       &0.992       &0.998       &0.023                                   &0.021       &0.010       \\
                                   & ToM    &\textbf{0.859}       &\textbf{0.994}       &\textbf{0.999}       &\textbf{0.027}       &\textbf{0.020}       &\textbf{0.011}       \\
                                   & Other  &0.899       &0.994       &0.999       &0.022       &0.021       &0.010      \\
        \bottomrule
    \end{tabular}
\end{table}

\subsection{Implementation details}
We use the Depth Anything V2 \cite{da2}, Depth Anything V1 \cite{da1}, Metric3D \cite{m3d} and DepthFM \cite{depthfm} as the baselines and fine-tune the Depth Anything V2 Large on the Hypersim dataset \cite{Hypersim}. 
We use the Hypersim dataset with random lighting conditions at a resolution of $518$ in both length and width for training. The total loss is $\mathcal{L}_{ToM}+\mathcal{L}_{ssi}$, in which $\mathcal{L}_{ssi}$ is the scale-shift-invariant loss function following MiDaS \cite{midas}. The max number of epochs is 5, the learning rate is set to 0.000005 and the batch size is 4. We also randomly flip the input samples with a 50\% probability during training. We also use a learning rate scheduling strategy with polynomial decay.

\subsection{Zero-shot Evaluations}

After the training on 70K purely synthetic samples has converged, our model demonstrates a strong zero-shot generalization capability. The quantitative comparison of our model on the NYU Depth Dataset V2 \cite{nyudv2} and Booster dataset \cite{booster} with the baselines are shown in Tab. \ref{tab:zero_shot_nyu} and Tab. \ref{tab:zero_shot_booster} respectively.

\begin{figure}[ht]
  \centering
  \includegraphics[height=6cm]{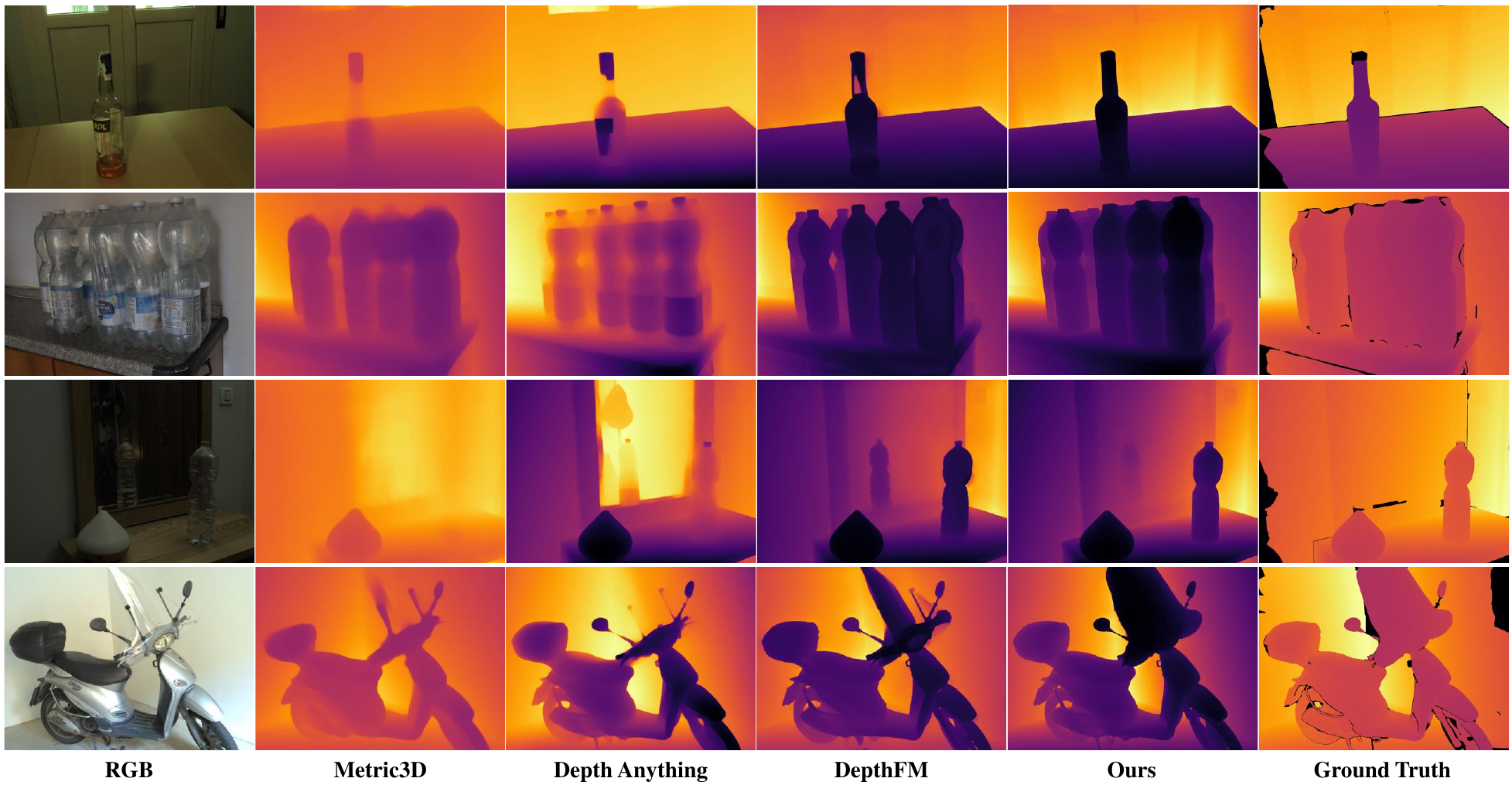}
  \caption{\textbf{Visual comparison of other baselines and our method.} By conducting zero-shot evaluations between other baselines and ours on the Booster dataset, results show that our method outperforms the baselines in most cases, as we explicitly and directly use non-Lambertian regional guidance.
  }
  \label{fig:others}
\end{figure}

The qualitative zero-shot evaluations on the Booster dataset and the NYU Depth Dataset V2 are show in Fig. \ref{fig:abstract} and Fig. \ref{fig:nyu} respectively. We also test the current state-of-the-art MDE method on the Booster dataset as shown in Fig. \ref{fig:others}. Our method exhibits a more reasonable overall depth structure and significantly better robustness to non-Lambertian surfaces compared to other methods. 

\begin{table}[ht]
    \centering
    \caption{Ablation studies of random tone-mapping augmentation (RTA) and non-Lambertian surface regional guidance (NSRG) during training and diverse lighting images fusion (DLIF) during testing. We conduct the evaluations on the Booster dataset. The best results in ToM regions are in \textbf{bold} and the second best results are \underline{underlined}.}
    \label{tab:ablation}
    \scalebox{0.9}{
    \begin{tabular}{cccccccccc}
        \toprule
        w/ RTA & w/ NSRG & w/ DLIF & Region & $\delta_{1.05}\uparrow$ & $\delta_{1.15}\uparrow$ & $\delta_{1.25}\uparrow$ & Abs Rel.$\downarrow$ & RMSE$\downarrow$ & Log MAE$\downarrow$ \\
        \midrule
        \multirow{3}{*}{}        & \multirow{3}{*}{}                 & \multirow{3}{*}{}              & All    & 0.802 & 0.975 & 0.993                             & 0.034 & 0.031 & 0.014                     \\
                                &                                   &                                & ToM    & 0.644 & 0.869 & 0.937 & 0.081 & 0.060 & 0.031                     \\
                                &                                   &                                & Other  & 0.830 & 0.988 & 0.998 & 0.029 & 0.025 & 0.013                   \\
        \midrule
        \multirow{3}{*}{\checkmark}        & \multirow{3}{*}{}                 & \multirow{3}{*}{}              & All    & 0.866                        &0.988                         &0.998                         &0.027                      &0.024                  &0.012                     \\
                                &                                   &                                & ToM    &0.800                         &0.984                         &\underline{0.998}                         &\underline{0.033}                      &\underline{0.023}                  &\underline{0.014}                     \\
                                &                                   &                                & Other  &0.876                         &0.991                         &0.998                         & 0.025                     &0.023                  &  0.011                   \\
        \midrule
        \multirow{3}{*}{}               & \multirow{3}{*}{\checkmark}                 & \multirow{3}{*}{}       & All    &0.865                         &0.988                         &0.997                         &0.026                      &0.024                  & 0.011                    \\
                                &                                   &                                & ToM    &0.815                         &0.986                         &\textbf{0.999}                         &\underline{0.033}                      &\underline{0.023}                  &\underline{0.014}                     \\
                                &                                   &                                & Other  &0.871                         &0.989                         &0.998                         &0.025                      & 0.023                 & 0.011                    \\
        \midrule
        \multirow{3}{*}{\checkmark}        & \multirow{3}{*}{\checkmark}          & \multirow{3}{*}{}       & All    &0.890       &0.992       &0.998       &0.023                                   &0.021       &0.010                     \\
                                &                                   &                                & ToM    &\underline{0.859}       &\underline{0.994}       &\textbf{0.999}       &\textbf{0.027}       &\textbf{0.020}       &\textbf{0.011}                     \\
                                &                                   &                                & Other   &0.880       &0.993       &0.999       &0.023       &0.022       &0.010                    \\
        \midrule
        \multirow{3}{*}{\checkmark}        & \multirow{3}{*}{\checkmark}          & \multirow{3}{*}{\checkmark}       & All    &0.885       &0.989       &0.997       &0.025                                   &0.022       &0.011                     \\
                                &                                   &                                & ToM    &\textbf{0.860}       &\textbf{0.997}       &\textbf{0.999}       &\textbf{0.027}       &\textbf{0.020}       &\textbf{0.011}                     \\
                                &                                   &                                & Other  &0.655       &0.913       &0.958       &0.059       &0.209       &0.025                    \\
        \bottomrule
    \end{tabular}}
\end{table}

While other models are trained on extensive datasets and leverage the semantic priors to guide the training, achieving good results on ToM regions, our model directly use the semantics of ToM regions to guide the model to have a more precise estimation for the depth in ToM regions. 

Our method achieves maximum accuracy improvements of 222.93\% and 192.75\% in the ToM regions on the Booster dataset and NYU deep dataset V2, respectively, compared to the baselines, demonstrating its strong generalization ability.

Since multi-exposure images of the same scene are available for Booster dataset, we also conduct diverse lighting images fusion during testing. More details are shown in Sec. \ref{section:fusion}.

\subsection{Ablation Study}
\label{section:ablation}

\begin{figure}[ht]
  \centering
  \includegraphics[height=5cm]{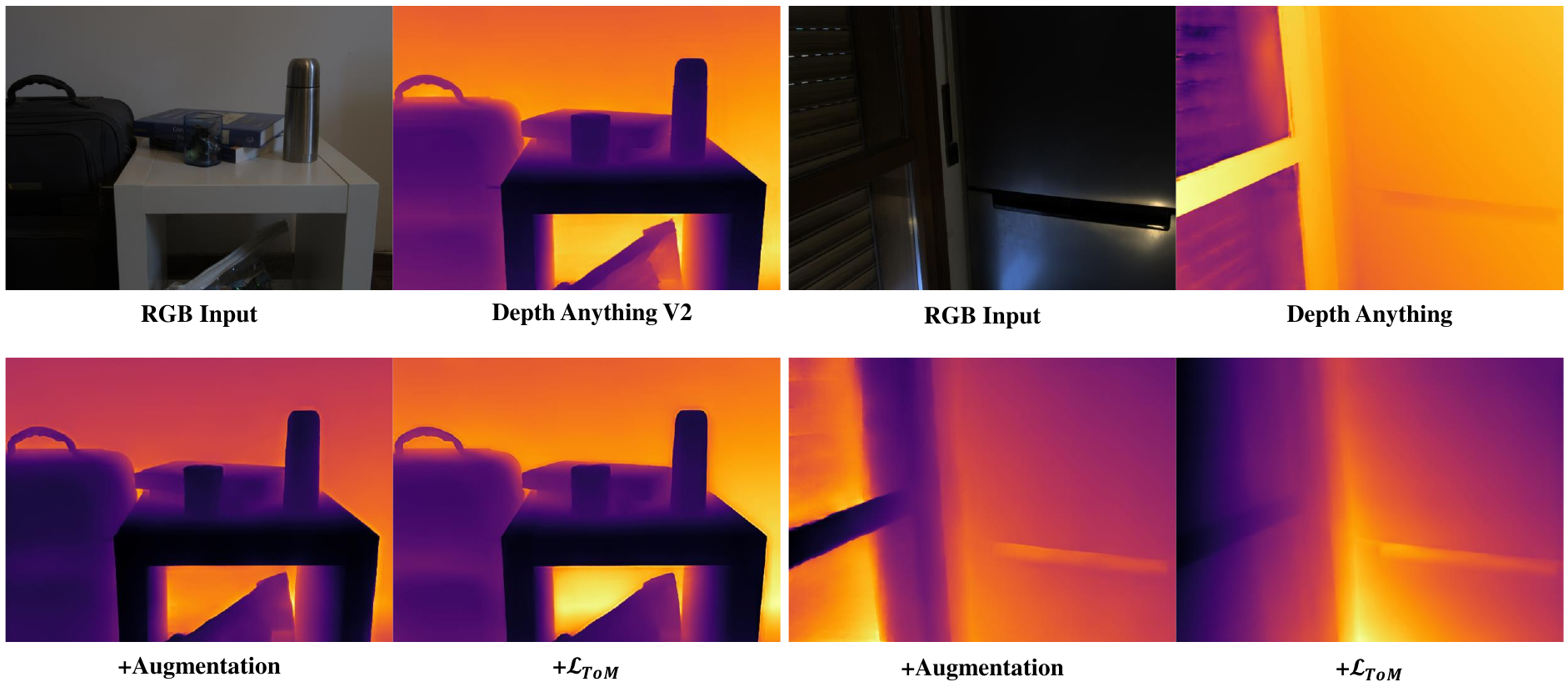}
  \caption{\textbf{Visual results of ablation study on random tone-mapping augmentation and non-Lambertian surface regional guidance.} Results consist of two scenes. For each scene, the RGB image is placed in the upper left, the depth estimated by the baseline is placed in the upper right, the depth estimated by our model trained using only random tone-mapping augmentation is placed in the lower left, and the depth estimated by our model trained using random tone-mapping augmentation and ToM surface regional guidance is placed in the lower right.
  }
  \label{fig:ablation}
\end{figure}

\subsubsection{Random Tone-mapping Augmentation} We first study the effectiveness of random tone-mapping augmentation on the Hypersim dataset during training. By adjusting a few random percentile values, we re-train the Depth Anything V2 on the Hypersim dataset with the new lighting conditions using the scale- and shift-invariant loss of MiDaS. In this way we expect the network to accommodate varying lighting conditions. As the visual results shown in Fig. \ref{fig:ablation}, our method can improve the performance of baseline in some regions with bad lighting conditions. At the same time, the $\delta_{1.05}$ accuracy of depth estimation in ToM regions gets 24.22\% improvement, shown in Tab. \ref{tab:ablation}.

\subsubsection{Non-Lambertian Surface Regional Guidance} Next, we train the baseline on the Hypersim dataset using $\mathcal{L}_{ToM}+\mathcal{L}_{ssi}$ as the loss function. As shown in Fig. \ref{fig:ablation} and Tab. \ref{tab:ablation}, by applying the non-Lambertian surface regional guidance, the model can more accurately predict the depth of non-Lambertian surfaces both qualitatively and quantitatively. The $\delta_{1.05}$ accuracy gets 26.55\% improvement in ToM regions compared to the baseline.

\subsubsection{Diverse Lighting Images Fusion}
\label{section:fusion}

\begin{figure}[H]
  \centering
  \includegraphics[height=6cm]{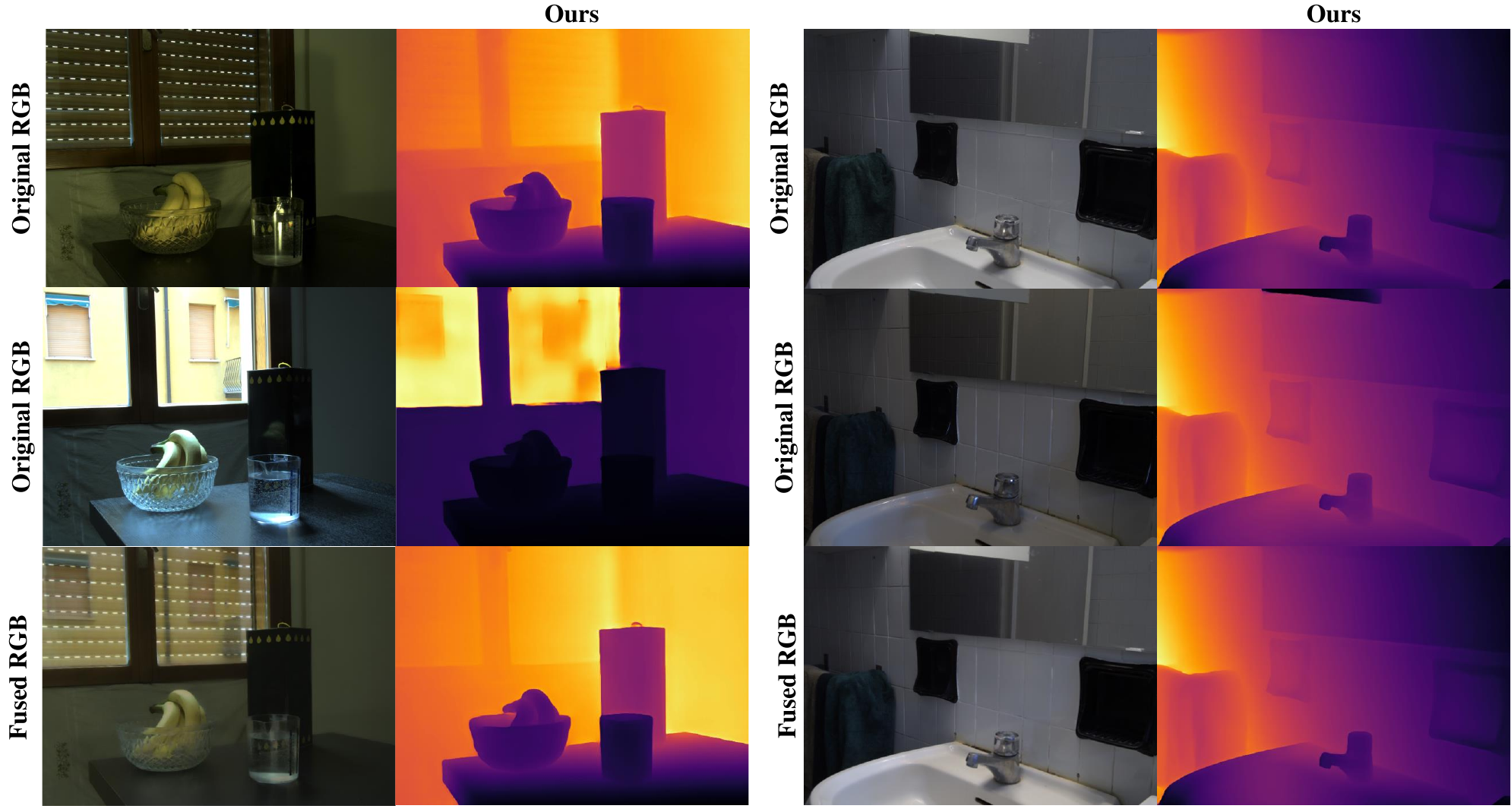}
  \caption{\textbf{Visual results of ablation study on diverse lighting images fusion.} The first and second rows are the results of our model trained using random tone-mapping augmentation and non-Lambertian surface regional guidance. The third row is the results of images fusion and the depth estimation of the fused image.
  }
  \label{fig:fuse}
\end{figure}

Although the model can learn to estimate the depth in different lighting conditions after training with random tone-mapping augmentation, in the testing phase, there are still some images with the same scene contents but different lighting conditions that may be predicted to have different depth. We propose to fuse the images using VAE to get a image with probably the best lighting conditions for the depth estimation. By inputting the fused RGB image, the predictions of our model are qualitatively superior to those obtained from directly inputting a single RGB image as shown in Fig. \ref{fig:fuse}. 
By combining multi-exposure images, the depth estimation gets improved as shown in Tab. \ref{tab:ablation}.

\section{Conclusion}
We propose a method to improve the performance of MDE models in the non-Lambertian regions and different lighting conditions. 
We encourage the network to directly learn the unique characteristics of non-Lambertian surfaces in depth prediction through a carefully designed training framework, rather than relying on in-painting of RGB images as in previous works.
After training on the synthetic dataset with non-Lambertian masks with \aug{} and \guidance{}, our model obtains a strong zero-shot capability. Moreover, in the testing phase, fusing images with the same scene contents but various lighting conditions can get a image with advantageous lighting conditions for depth estimation, contributing to more accurate prediction results especially in non-Lambertian regions. However, our work still has some limitations. For example, in cases of extremely overexposed images with poor quality, the textures of non-Lambertian surfaces like glass are almost completely missing, making it possible for the network to fail in its predictions. Future work can better integrate image quality optimization and semantic information fusion into the depth estimation of non-Lambertian surfaces, achieving better qualitative and quantitative results.

%
%
\bibliographystyle{splncs04}
\bibliography{main}
\end{document}